\documentclass[letterpaper, 10 pt, conference]{ieeeconf}  

\IEEEoverridecommandlockouts                              

\usepackage{amsmath, amssymb , graphicx, subcaption}
\usepackage{multirow}
\usepackage{dblfloatfix} 
\usepackage{balance}

\DeclareCaptionFont{caption}{\fontsize{8}{9.6}\selectfont}
\title{\LARGE \bf 
Motion Scaling Solutions for Improved Performance in High Delay Surgical Teleoperation
}

\author{Florian Richter$^1$, Ryan K. Orosco$^2$ \IEEEmembership{Member, IEEE}, and Michael C. Yip$^1$ \IEEEmembership{Member, IEEE}
\thanks{$^1$Florian Richter and Michael C. Yip are with the Department of Electrical and Computer Engineering, University of California San Diego, La Jolla, CA 92093 USA. {\tt\small \{frichter, yip\}@ucsd.edu}}%
\thanks{$^2$Ryan K. Orosco is with the Department of Surgery - Division of Head and Neck Surgery, University of California San Diego, La Jolla, CA 92093 USA. {\tt\small rorosco@ucsd.edu}}}

\begin{document}

\maketitle
\thispagestyle{empty}
\pagestyle{empty}

\begin{abstract}
Robotic teleoperation brings great potential for advances within the field of surgery. The ability of a surgeon to reach patient remotely opens exciting opportunities. Early experience with telerobotic surgery has been interesting, but the clinical feasibility remains out of reach, largely due to the deleterious effects of communication delays. Teleoperation tasks are significantly impacted by unavoidable signal latency, which directly results in slower operations, less precision in movements, and increased human errors. Introducing significant changes to the surgical workflow, for example by introducing semi-automation or self-correction, present too significant a technological and ethical burden for commercial surgical robotic systems to adopt. In this paper, we present three simple and intuitive motion scaling solutions to combat teleoperated robotic systems under delay and help improve operator accuracy. Motion scaling offers potentially improved user performance and reduction in errors with minimal change to the underlying teleoperation architecture. To validate the use of motion scaling as a performance enhancer in telesurgery, we conducted a user study with 17 participants, and our results show that the proposed solutions do indeed reduce the error rate when operating under high delay.
\end{abstract}

\section{Introduction}

Teleoperational control allows the operator to complete tasks at a safe, remote location. This is accomplished by having a human operator send control signals from a remote console, traditionally known as \textit{master}, to the robot performing the task, traditionally known as \textit{slave}. The slave then returns feedback to the master through sensors, such as cameras and haptic devices, which gives the operator the feeling of \textit{telepresence} through displays and force feedback. Through improvement of telepresence and robotic systems, teleoperational systems are becoming common place in a wide range of applications such as underwater exploration, space robotics, mobile robots, and telesurgery \cite{TeleoperationHistory, yipDasJournal} . 

With the advent of surgical robots such as the da Vinci\textregistered{} Surgical System, attempts have been made to investigate the feasibility of remote telesurgery. Anvari et al. in 2005 was first to establish a remote telesurgical service and reported on 21 remote laparoscopic surgeries with an observed delay of 135-140 msec over the 400 km \cite{firstTeleRobotic}. The signal latency grows as the distance increases, and prior reports have cited 300 msec as the maximum time delay where surgeons began to consider the operation unsafe \cite{TransatlanticSurgery, LatencyEffects}. When using satellite communication between London and Toronto, a report measured a delay of $560.7 \pm 16.5$ msec \cite{reasonForDelay}. This unavoidable time delay is the greatest obstacle for safe and effective remote telesurgery because it leads to overshoot and oscillations. These undesirable effects have been observed with as little as 50 msec of time delay in surgical tasks \cite{SteadyHand, TreatmentPlanning}. The challenges of teleoperating under delay has been investigated for over 50 years, and the field is too expansive to cover in this paper. A more broad look can be found in Hokayem and Spong's historical survey \cite{TeleoperationHistory}.

The first study on the effects of delay in teleoperational system were by Ferrell in 1965 and found that experienced operators will use a strategy called move and wait \cite{MoveAndWait}. A stable system is realized by inputting a new control command and then waiting to see the effect. Ferrell and Sheridan then developed supervisory control to address the problem of delay \cite{supervisoryControl}. This gives a level of autonomy to the system such that the operator supervises rather than explicitly inputting trajectory motions. Therefore, supervisory control can only be implemented practically at this time in structured environments.

In the case of teleoperating with haptic feedback, also known as \textit{bilateral} teleoperation, delay has been experimentally and theoretically shown to create instability within the system \cite{experimentalInstability, theoryInstability}. Techniques such as wave variables have been used to dampen the unstable overshoot in bilateral control under delay \cite{waveVariables}. However, they have been shown to increase time to complete task \cite{yipBadHaptic, yipBadHaptic_2}, and therefore teleoperating without haptic feedback can often be a better alternative.

Predictive displays circumvent the deleterious effects from teleoperating under delay by giving the operator immediate feedback through a virtual prediction. Early predictive displays focused on space robotics where delays can reach up to 7sec \cite{spaceTeleopDelay2} \cite{spaceTeleopDelay3}. Winck et al. recently applied this approach by creating an entire predicted virtual environment that is displayed to the operator and is also used to generate haptic feedback \cite{spaceTeleopDelay}. Currently this is not applicable to telesurgery because of the unique 3D geometry found in a surgical environment. The scale of the operation requires high precision tracking to create reliable predictions, and obstacles such as tissue cannot be modeled as rigid bodies nor accurately predicted.

In this paper, we present new motion scaling solutions for conducting telesurgery under delay that:
\begin{enumerate}
	\item reduce errors to improve patient safety,
	\item have intuitive control since the delay can change from one operation to another due to variable distances, and
	\item have simple implementation so it can be easily deployed to any teleoperated system.
\end{enumerate}

To show the performance of the solutions with statistical significance, we conducted a user study on the da Vinci\textregistered{} Surgical System. The user study was designed such that there is independence from unwanted variables such as participant exhaustion and experience and that the best performing solution has a short learning curve, therefore intuitive. Both time to complete task and an error metric were used to evaluate task performance, and we measured that our proposed solutions decreased the number of errors at the cost of time. In fact, 16 out of the 17 participants from the user study performed best, with regards to error rate, using our proposed solutions.

\section{Methods} 

\begin{figure}[b]
	\centering
	\includegraphics[width=0.49\textwidth]{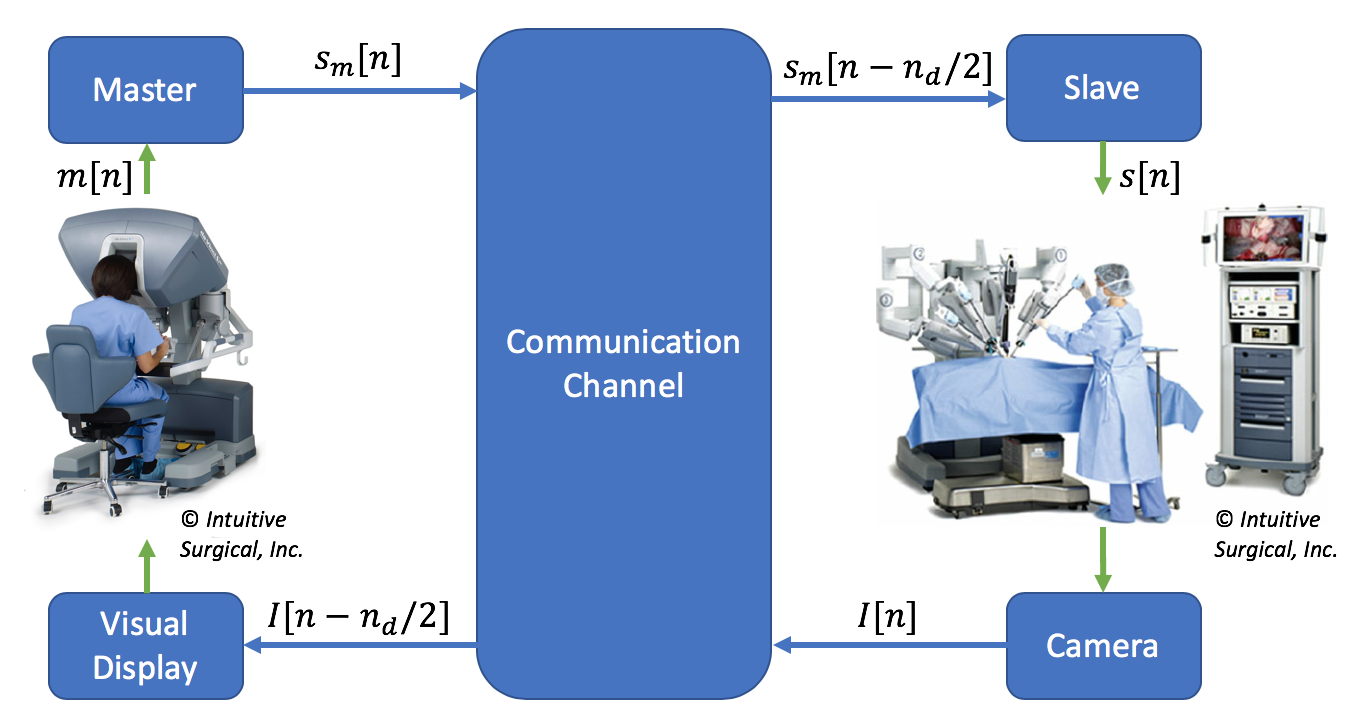}
	\caption{Flowchart of a teleoperation system with round trip delay of $d$ where $m$ is the masters position, $s_m$ is the target slave position from the master, $s$ is the slave position, $I$ is feedback to the operator. For the sake of simplicity, the entire system is assumed to have a sampling rate of $f_s$, and let $n_d = f_sd$ }
	\label{fig:flowChart}
\end{figure}

Teleoperation systems utilizes a scaling factor   between master arms movements and the slave arms movements to adapt the teleoperator to the slave workspace. It can be expressed as a general teleoperation system with delay, shown in Fig. \ref{fig:flowChart}, as:
\begin{equation}
	s_m[n] = scale_m(m[n] - m[n-1]) + s_m[n-1]
\end{equation}
\begin{equation}
	s[n] = s_m[n-n_d/2]
\end{equation}
where $scale_m$ is the scaling factor previously described. It is worth noting that teleoperation systems described in this manner using position control, such as the da Vinci\textregistered{} Surgical System, only apply this scaling to position and not orientation, rather mapping orientation commands one-to-one so there is no obvious mismatch between user wrist orientation and robot manipulator orientations. The solutions we present are all modifications of Equations (1) and (2).

\subsection{Constant Scaling Solution}

The first solution simply reduces the scaling factor in Equation (1). Lowering the scaling factor in robotic surgery has been shown to increase task completion time and improve accuracy \cite{motionScaling1, motionScaling2}. We extend this to teleoperating under delay and hypothesize that decreasing the constant scale will reduce the number of errors at the cost of time when operating under delay. This idea is similar to the move and wait strategy, creating stability by slowing down the motions. However, here the motions are slowed continuously, while the move and wait strategy discretizes the motions and requires the operator to have experience working under the delay. 

\subsection{Positional Scaling Solution}

Positional scaling builds on the constant scaling solution by decreasing the scaling as the slave arms move towards an obstacle. Therefore, reducing the cost of time from the constant scaling solution by slowing down the operators motions only in areas of the work space that require higher precision. Since this deals with obstacle detection, the positional scaling will be implemented on the slave side through the following equations:
\begin{equation}
	scale_s = \text{min}(maxscale, \text{max}(minscale, k*r[n]))
\end{equation}
\begin{equation}
	s[n] = scale_s(s_m[n-n_d/2] - s_m[n-n_d/2-1]) + s[n-1]
\end{equation}
where $r$ is the distance from $s_m$ to the nearest obstacle, $k$ is the rate $scale_s$ changes, and $minscale$ and $maxscale$ set the lower and upper bounds respectively of $scale_s$. This is paired with Equation (1), and can be thought of as secondary layer of scaling based on proximity to an obstacle.

To implement this scaling method, the distance $r$ requires the location of the obstacles in the environment. Methods such as feature-based 3D tissue reconstruction from \cite{tissue3DReconstruction} or dense pixel-based depth reconstruction can be used. In the presented user study, the task environment is known and distances $r$ are found analytically. 

\begin{figure*}[t]
	\centering
	\begin{subfigure}{.24\textwidth}
  		\centering
  		\includegraphics[width=1\linewidth]{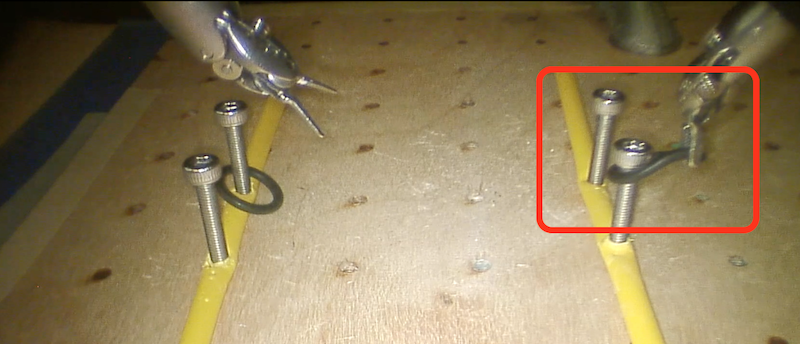}
	\end{subfigure}
	\begin{subfigure}{.24\textwidth}
		\centering
  		\includegraphics[width=1\linewidth]{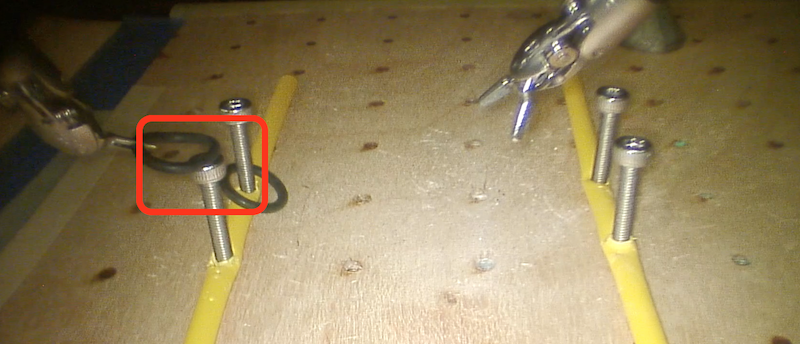}
	\end{subfigure}
	\begin{subfigure}{.24\textwidth}
  		\centering
  		\includegraphics[width=1\linewidth]{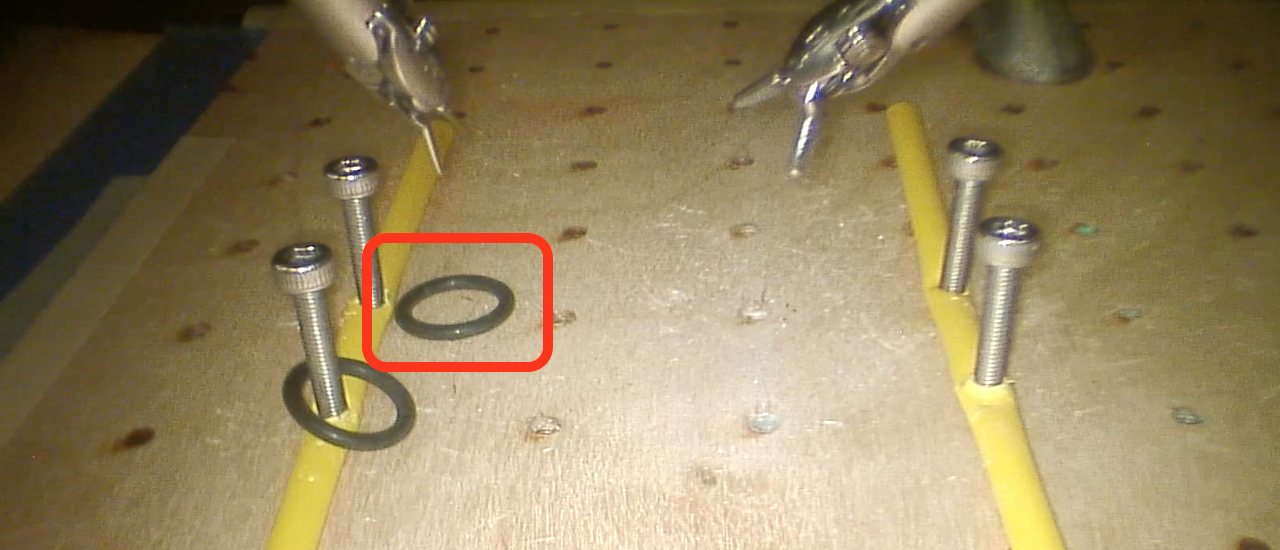}
	\end{subfigure}
	\begin{subfigure}{.24\textwidth}
		\centering
  		\includegraphics[width=1\linewidth]{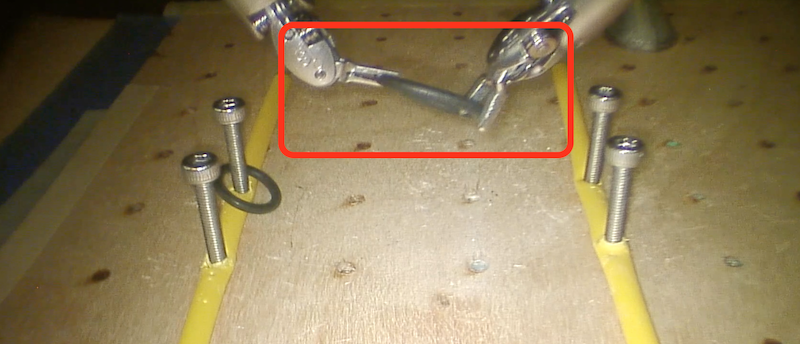}
	\end{subfigure}
	\caption{Photos of example errors from the user study. From left to right the errors are: stretch ring on peg, touch peg, drop ring, and stretch ring during handoff.}
	\label{fig:sampleErrors}
\end{figure*}

\subsection{Velocity Scaling Solution}

Velocity scaling is founded on the idea that an operator will naturally move slower when higher accuracy is required. This is inspired by the software controlling computer mice and trackpads, which natively utilize this strategy by default (unless turned off by the owner). Combining velocity scaling with the constant scaling solutions hypothesis, the scaling factor in Equation (1) is increased proportionally to the input translational velocity as shown:
\begin{equation}
	scale_m = v_1 + v_2 \big| \big| \dot{m}_m[n] \big| \big|
\end{equation}
where $v_1$ is the base scaling, the bonus velocity scaling is given at the rate of $v_2$, $\dot{m}$ is the translational velocity, and $|| \cdot ||$ is the magnitude. Equation (2) is then used to set the slave position on the slave side. This allows for a low base scaling to get the benefits of the constant scaling solution when the operator is making small, precise motions. It also increases the scaling when the operator makes large motions, which requires less precision, in order to reduce the cost of time.

To find $\dot{m}$ on a real system where there is noise, the following filter can be used: 
\begin{equation}
	\dot{m}_m[n] = \frac{m[n] + m[n-1] - m[n-2] - m[n-3]}{4/f_s} 
\end{equation}\\
This represents a running average with weights 0.25, 0.5, and 0.25 for three individual velocity measurements.

\section{Experiment}

To measure the effectiveness of the solutions, a user study with 17 participants was conducted on the da Vinci\textregistered{} Surgical System. The delayed feedback to the user is a stereoscopic 1080p laparoscopic camera running at 30FPS. The camera feed is displayed to a console for stereo viewing by the operator. Both pairs of master and slave arms have seven degrees of freedom. A modified version of Open-Source da Vinci Research Kit \cite{DVRK} was used on an a computer with an Intel\textregistered{} Core\texttrademark{} i9-7940X Processor and NVIDIA's GeForce GTX 1060. 


To ensure that the solutions are tested in a realistic, high delay condition, a round trip delay of 750 msec was simulated for the delayed environment in the user study. This follows a  recent report showing satellite communication between London and Toronto measured a round trip delay of $560.7 \pm 16.5$ msec for a telesurgical task \cite{reasonForDelay}. 

\subsection{Task}
\begin{figure}[b]
	\centering
	\includegraphics[width=0.48\textwidth]{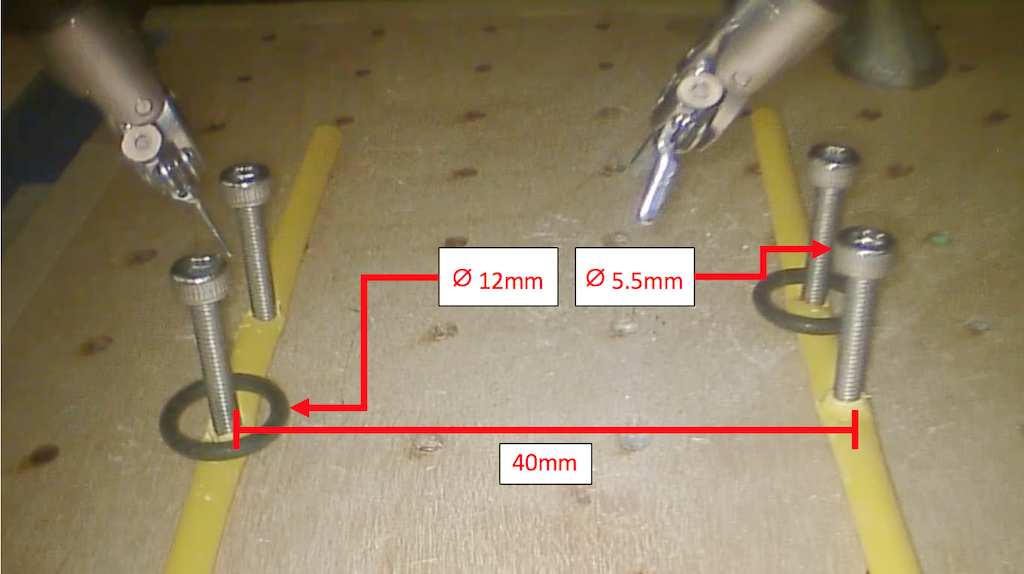}
	\caption{Task environment for the user study as seen from the endoscope. The scale of the environment ensures that participants must be accurate in order to perform well with regards to errors.}
	\label{fig:taskEnv}
\end{figure}

A peg transfer task is used as a test scenario. Fig. \ref{fig:taskEnv} shows a photo from the endoscope used by the operator in our study. The task involves:

\begin{enumerate}
\item Lifting a rubber o-ring from either the front left or right peg with the corresponding left or right arm
\item Passing the ring to the other arm
\item Placing the ring on the other front peg
\item Repeating steps 1 to 3 for the back pair of pegs
\end{enumerate}

This task and environment was chosen because it inherits complex motions such as hand off, regions where larger movements are safe, and yet still requires precise motions during parts of the trajectory due to the tightness of fit between the rings and the pegs.

\subsection{Metrics}

The metrics to evaluate task performance are time to complete task and an error metric. The error metric is an enumeration of different types of errors that we observed. These errors are weighted according to Table 1 and summed to get a weighted error. The weights were chosen such that severity of the error would be reflected properly in the error metric. Example errors are shown in Fig. \ref{fig:sampleErrors}. A sample video was created and shown to the participants before the study to show the task and the different types of errors.

\begin{table}[h]
\caption{\\Weights associated with type of error}
\begin{center}
\begin{tabular}{ |c|c| } 
\hline  
	\textbf{Error} & \textbf{Weight}								\\   \hline
	Touch peg & 1				 							\\   \hline
	Touch ground & 2					 					\\   \hline
	Stretch ring during hand-off for a second or less & 2 			\\   \hline
	Drop ring & 3					 						\\   \hline
	Stretch ring on peg for a second or less & 4					\\   \hline
	Stretch ring for an additional second & 4					 	\\   \hline
	Stretch/move peg & 10					 				\\   \hline
	Knock down peg & 20					 				\\   \hline
\end{tabular}
\end{center}
\end{table}

\subsection{Procedure}
The scaling scenarios for the participants to complete the peg transfer task are as follows in both 0  msec and 750  msec round trip delay:
\begin{enumerate}
	\item Constant scaling of 0.3
	\item Constant scaling of 0.2
	\item Constant scaling of 0.1
	\item Position scaling
	\item Velocity scaling
\end{enumerate}

For constant scaling, the corresponding scaling value listed above was used for $scale_m$. For positional scaling, the following simplification was made: $r$ is the minimum 2D distance from center of each of the 4 pegs to the target tool-tip position projected on the plane constructed from the top of the 4 pegs ($ax + by + z = c$). The following equations are used to find the least squared solution of the plane:

\begin{equation}
    A = 
\begin{bmatrix}
x_1 & y_1 & -1 \\
x_2 & y_2 & -1 \\
x_3 & y_3 & -1 \\
x_4 & y_4 & -1 \\
\end{bmatrix}
\end{equation}
\begin{equation}
    \begin{bmatrix}
        a & b & c
    \end{bmatrix}^T
    = -(A^T A)^{-1}A^T
    \begin{bmatrix}
        z_1 & z_2 & z_3 & z_4
    \end{bmatrix}^T
\end{equation}
where $\begin{bmatrix} x_i & y_i & z_i \end{bmatrix}^T$ is the position of the center of the $i$-th peg. The following is computed realtime to project the target tool-tip position and find $r$:
\begin{equation}
    e = \frac{\begin{bmatrix} a & b & 1 \end{bmatrix}^T}{\big|\big|\begin{bmatrix} a & b & 1 \end{bmatrix}^T\big|\big|}
\end{equation}
\begin{equation}
    s_p[n] = s_m[n-n_d/2] - \begin{bmatrix} 0 & 0 & c \end{bmatrix}^T - \text{ proj}_e s_m[n-n_d/2]
\end{equation}
\begin{equation}
    r[n] = \text{ }\underset{i}{\operatorname{min}}\text{ } \big|\big| s_p[n] - \begin{bmatrix} x_i & y_i & z_i \end{bmatrix}^T \big|\big|
\end{equation}

\begin{figure}[b]
	\centering
	\includegraphics[width=0.48\textwidth]{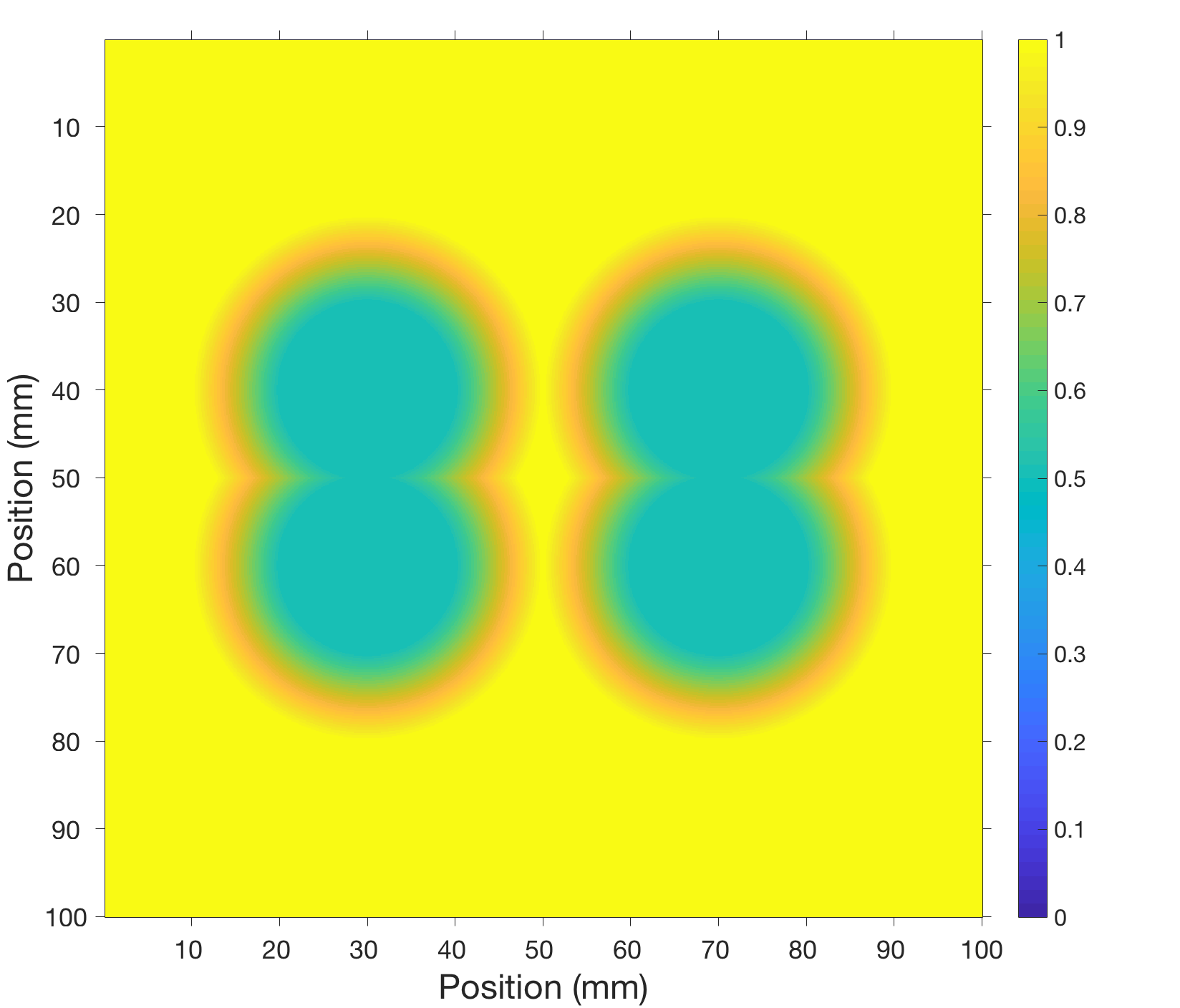}
	\caption{Map showing the value of $scale_s$ for the positional scaling used in the user study. This map is from a bird-eye view of the pegs which are located at: (30, 40), (30, 60), (70, 40), and (70, 60). }
	\label{fig:positionScalingPlot}
\end{figure}

Through experimentation, $scale_m = 0.2$, $k = 100$, $minscale= 0.5$, and $maxscale = 1.0$ were chosen for positional scaling. Fig. \ref{fig:positionScalingPlot} shows a generated map of the value of $scale_s$ using these parameters. $scale_s$ is calculated and applied to both arms individually. These positional scaling parameters mean that the total scaling from master to slave ranges from 0.1 to 0.2.

For velocity scaling, the values of $v_1 = 0.1$ and $v_2 = 100$ were found through experimentation. Similar to positional scaling, this is calculated and applied to both arms individually.

The procedure for each participant was as follows:
\begin{enumerate}
	\item Practice: repeat the task twice with constant scaling of 0.2 under no delay
	\item Record: perform the task with constant scaling of 0.2 under no delay
	\item Practice: select a new scaling method at random under no delay and take 20 seconds to become accustomed to the new scenario
	\item Record: perform the task with the previously selected scaling solution
	\item Repeat step 3 and 4 with all scaling solutions 
	\item Repeat step 3 and 4 with constant scaling of 0.2 
	\item Repeat 1-6 with round trip delay of 750 msec.
\end{enumerate}

Step 1 is so that participants get over the learning curve of the task and the system. By comparing the results of step 2 and 6, we can ensure that the user overcame the learning curve in step 1 and has not degraded in performance due to exhaustion. This is critical since it shows that the performance of a participant is only affected by the delay and different scaling methods. To further break any potential correlation, the randomization of the order of scaling methods in step 3 and 4 was used. Each participant has only 20 seconds to practice in the new scenario (step 3), so that the solutions with best results will be intuitive.

\section{Results}

To show there is no significant change in performance between step 2 and 6 of the procedure, the results are compared using two-sided paired-sample t-tests. Under no delay, $p = 0.670$ and $p = 0.054$ for weighted error and time to complete task respectively is computed. For the roundtrip delay of 750 msec, $p = 0.633$ and $p = 0.192$ for weighted error and time to complete task respectively is computed. Therefore, no statistically significant ($p<0.05$) change in performance was measured. Combining this result with the randomization of order in step 3 of the procedure implies that the measured results will be uncorrelated with participant experience and exhaustion and only affected by the delay and scaling solution.

\begin{figure*}[t]
	\centering
	\begin{subfigure}{.4\textwidth}
  		\centering
  		\includegraphics[width=1\linewidth]{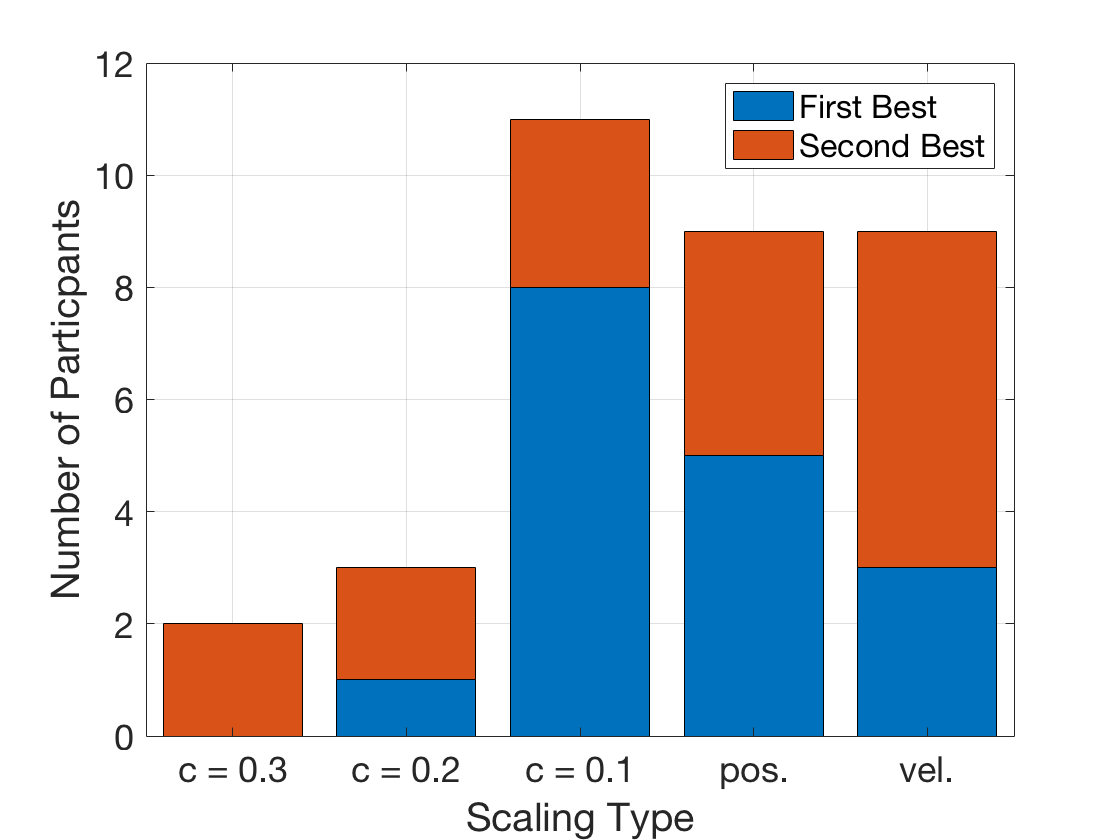}
	\end{subfigure}
	\qquad
	\qquad
	\begin{subfigure}{.4\textwidth}
		\centering
  		\includegraphics[width=1\linewidth]{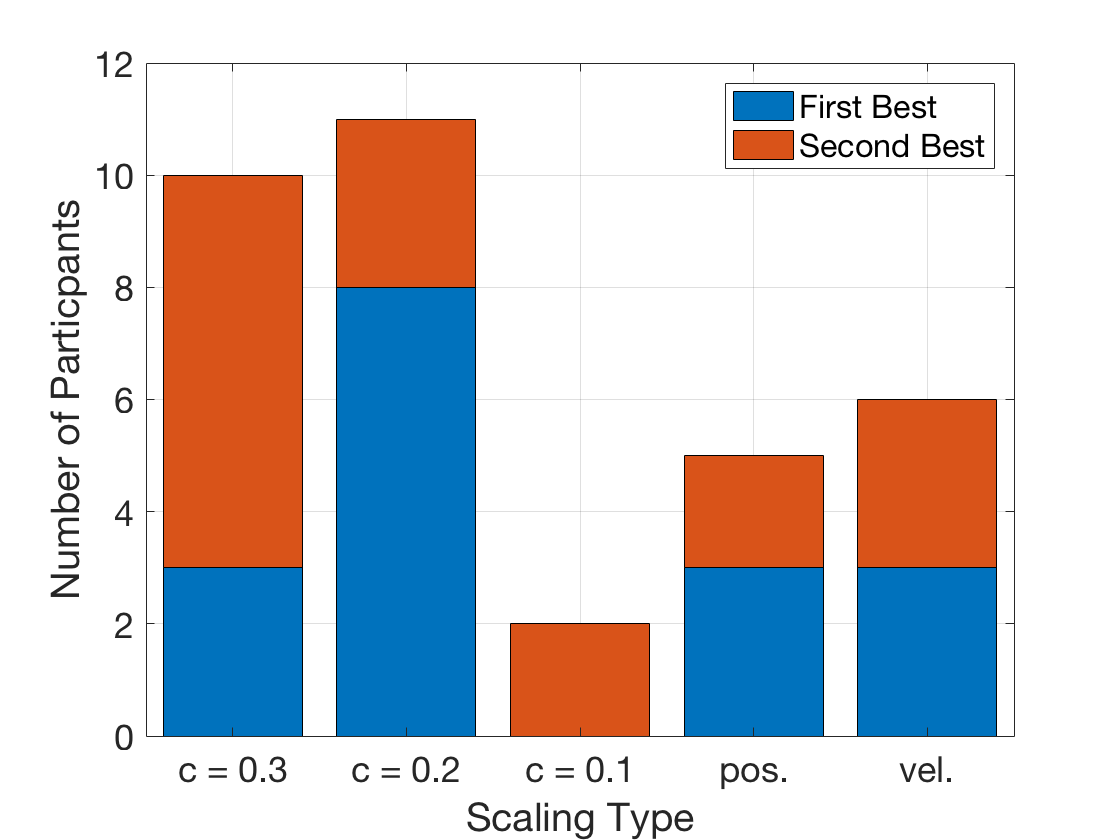}
	\end{subfigure}
	\caption{Counts of participants for which scaling solution performed best and second best from the user study under a roundtrip delay of 750 msec. Left is weighted error and right is time to compete task.}
	\label{fig:bestScaling}
\end{figure*}

The results showing which solution performed best and second best for each individual person under delay are shown in Fig. \ref{fig:bestScaling}. Only one out of seventeen participants performed best, with regards to weighted error, using the baseline scaling, constant scaling of 0.2 and 0.3. Furthermore, from Fig. \ref{fig:bestScaling} it is evident that even when including the second best performing solution, the vast majority of participants performed better with our proposed solutions with regards to weighted error. 

\begin{figure*}[t]
	\centering
	\begin{subfigure}{0.4\textwidth}
            \centering
            \includegraphics[width=\linewidth]{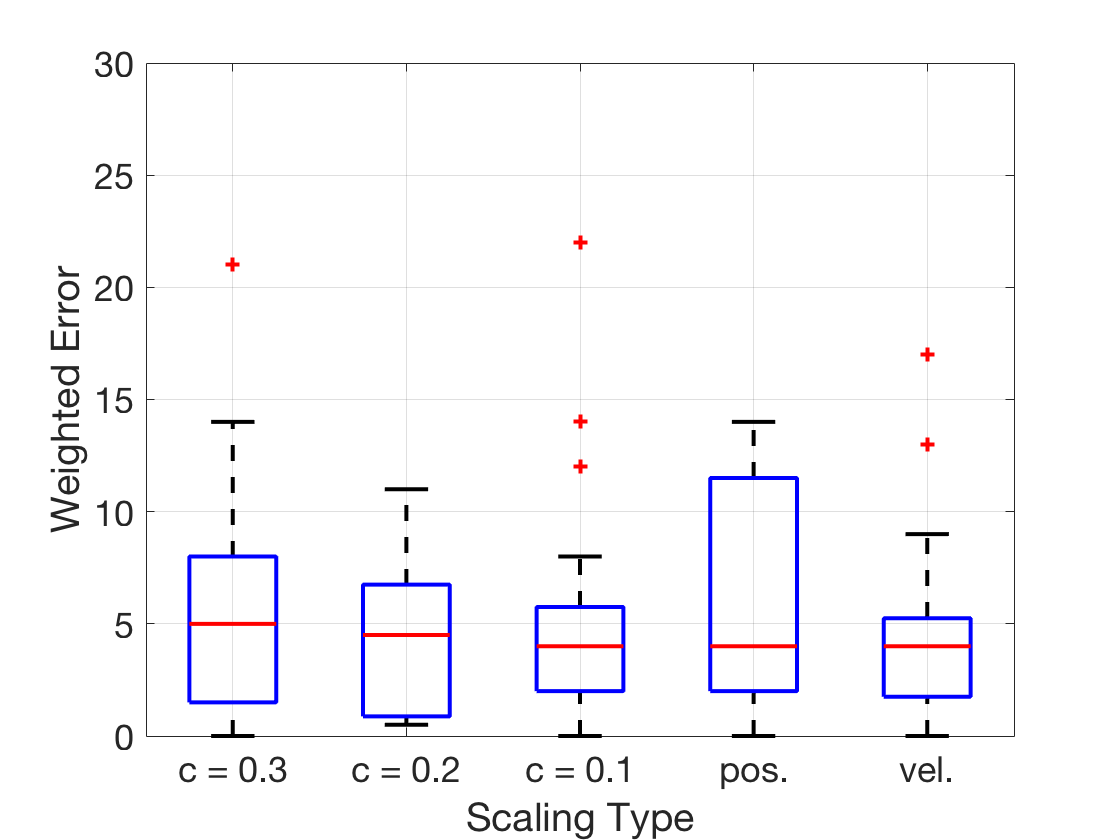}
        \end{subfigure} 
        \qquad\qquad
        \begin{subfigure}{0.4\textwidth}  
            \centering 
            \includegraphics[width=\linewidth]{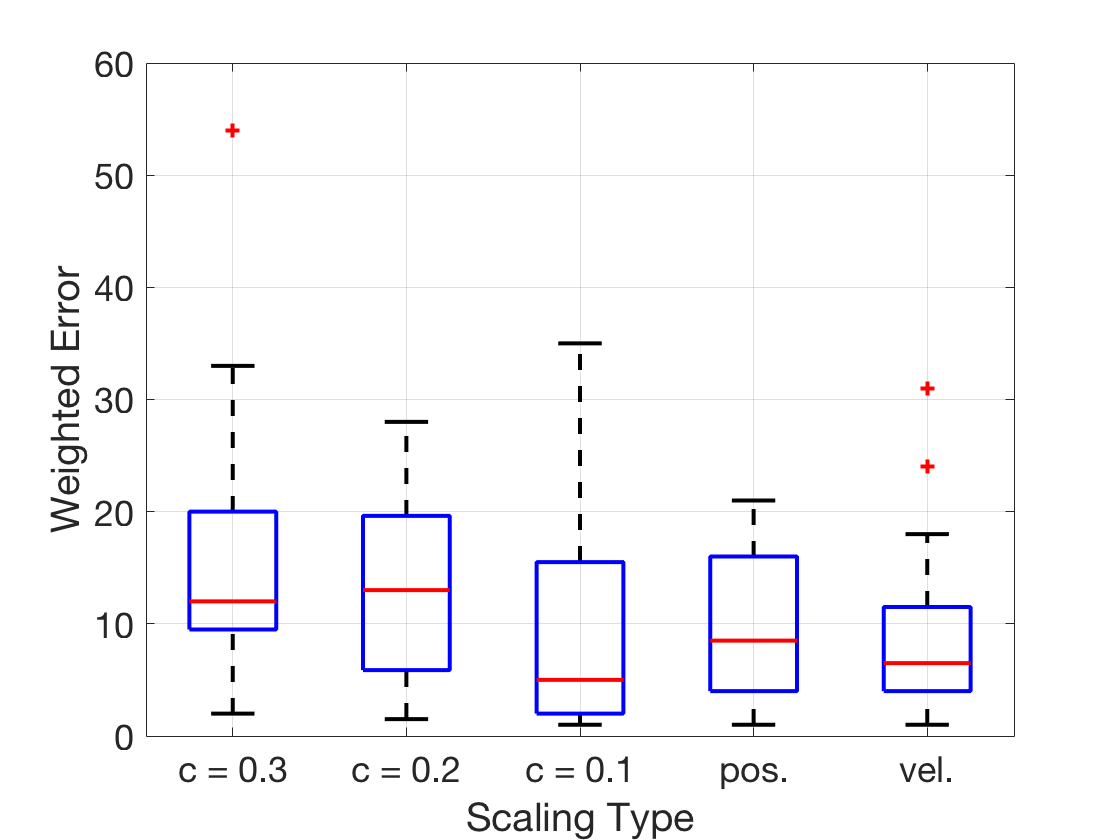}
        \end{subfigure}
        \vskip\baselineskip
        \begin{subfigure}{0.4\textwidth}   
            \centering 
            \includegraphics[width=\linewidth]{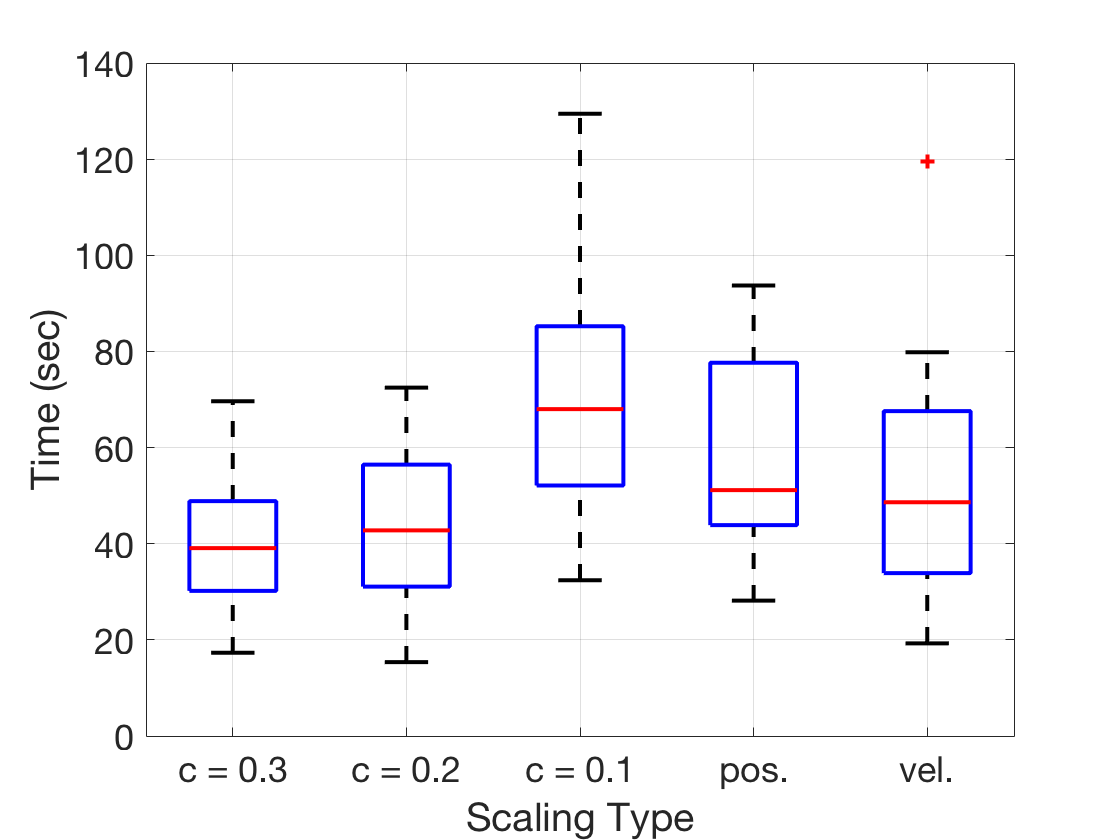}
        \end{subfigure}
        \qquad\qquad
        \begin{subfigure}{0.4\textwidth}   
            \centering 
            \includegraphics[width=\linewidth]{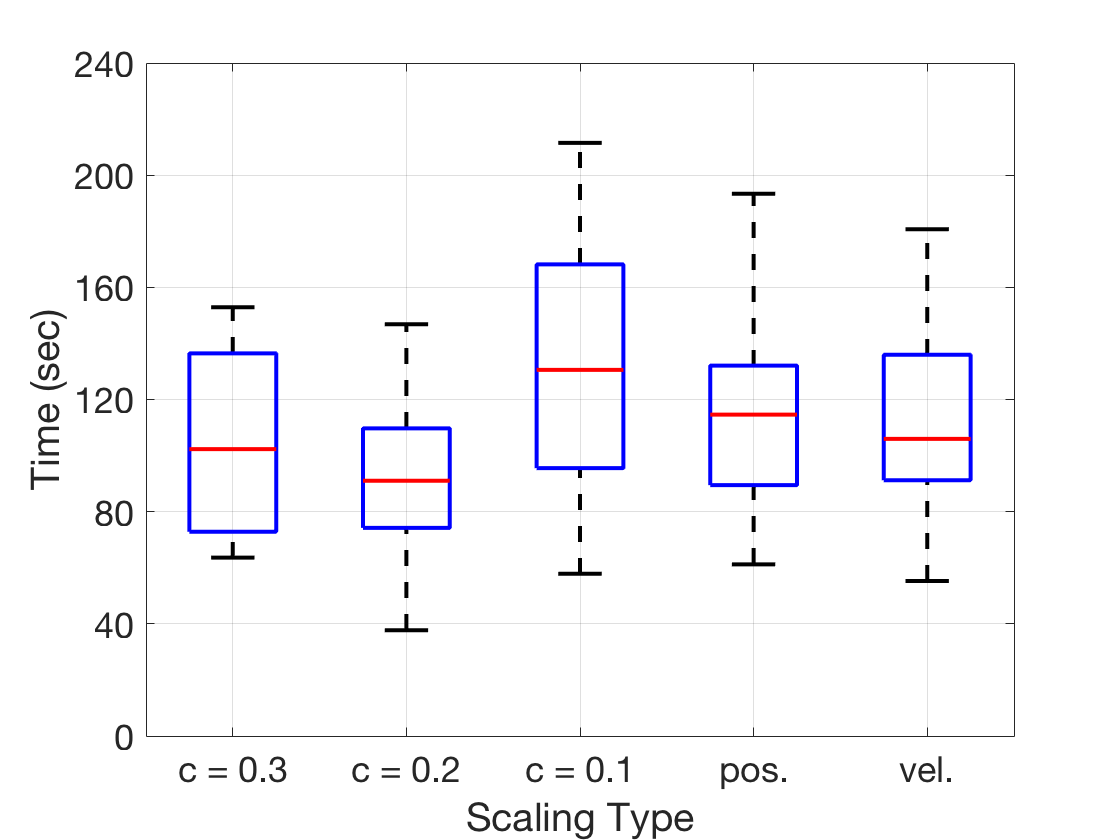}
        \end{subfigure}%
        \caption{Performance results of all tested scaling methods from the user study. Plots on the left are results from no delay and plots on the right are results with a roundtrip delay of 750 msec.} 
        \label{fig:allDataPlots}
\end{figure*}

The statistics of the results are shown in Table 2 and 3 for roundtrip delay of 750 msec and 0sec respectively. The $p$-values are generated by comparing against the baseline scaling with a two-sided paired t-test to show if the solution had a statistically significant ($p < 0.05$) effect on performance. The box plots of the results are shown at the end of the paper in Fig. \ref{fig:allDataPlots}. All the proposed solutions do achieve statistically significant lower weighted error under delay except for positional scaling compared against constant scaling of 0.2. The proposed solutions were also measured to make a statistically significant increase in time under delay when comparing against constant scaling of 0.2. Therefore, the solutions are working as intended, increasing the accuracy at the cost of time. 

\begin{table*}[t]
\centering
\caption{\\Statistics from user study trials under a roundtrip delay of 750 msec. $p$-value is generated from two-sided paired t-test.}
\def\arraystretch{1.25}
\setlength\tabcolsep{1.25em}
\begin{tabular}{c|ccc|ccc}
      & \multicolumn{3}{c|}{\textbf{Weighted Error}} & \multicolumn{3}{c}{\textbf{Time (sec)}} \\ 
         & $mean \pm std$    & $p$-value       & $p$-value       & $mean \pm std$    & $p$-value       & $p$-value \\ 
                &                   & vs. c = 0.3   & vs. c = 0.2    &                   & vs. c = 0.3   & vs. c = 0.2    \\\hline
    \textbf{c = 0.3} & 16.4 $\pm$ 13.3 & --- & 0.305 & 106 $\pm$ 32.0 & --- & 0.224 \\
    \textbf{c = 0.2} & 13.2 $\pm$ 8.22 & 0.305 & --- & 92.5 $\pm$ 31.8 & 0.224 & --- \\
    \textbf{c = 0.1} & 9.29 $\pm$ 9.62 & 0.0149 & 0.0480 & 131 $\pm$ 42.8 & 0.000316 & $<0.0001$ \\
    \textbf{pos.} & 12.3 $\pm$ 14.6 & 0.0245 & 0.572 & 116 $\pm$ 38.0 & 0.173 & 0.0474 \\
    \textbf{vel.} & 9.31 $\pm$ 8.46 & 0.00926 & 0.00114 & 113 $\pm$ 35.1 & 0.284 & 0.00617 \\
\end{tabular}
\end{table*}

\begin{table*}[t]
\centering
\caption{\\Statistics from user study trials under a roundtrip delay of 0sec. $p$-value is generated from two-sided paired t-test.}
\def\arraystretch{1.25}
\setlength\tabcolsep{1.25em}
\begin{tabular}{c|ccc|ccc}
      & \multicolumn{3}{c|}{\textbf{Weighted Error}} & \multicolumn{3}{c}{\textbf{Time (sec)}} \\ 
         & $mean \pm std$    & $p$-value       & $p$-value       & $mean \pm std$    & $p$-value       & $p$-value \\ 
                &                   & vs. c = 0.3   & vs. c = 0.2    &                   & vs. c = 0.3   & vs. c = 0.2    \\\hline
    \textbf{c = 0.3} & 6.00 $\pm$ 5.60 & --- & 0.266 & 40.3 $\pm$ 13.7 & --- & 0.320 \\
    \textbf{c = 0.2} & 4.47 $\pm$ 3.40 & 0.266 & --- & 44.3 $\pm$ 17.6 & 0.320 & --- \\
    \textbf{c = 0.1} & 5.35 $\pm$ 5.73 & 0.686 & 0.486 & 72.7 $\pm$ 27.9 & $<0.0001$ & $<0.0001$ \\
    \textbf{pos.} & 6.12 $\pm$ 5.01 & 0.833 & 0.243 & 60.0 $\pm$ 20.5 & 0.000246 & 0.000100 \\
    \textbf{vel.} & 4.76 $\pm$ 4.49 & 0.259 & 0.784 & 53.2 $\pm$ 24.5 & 0.0123 & 0.0965 \\
\end{tabular}
\end{table*}

\section{Discussion}

The user study results of constant scaling of 0.1 under delay match with our initial hypothesis that decreasing constant scaling slows down the operators motions to achieve higher accuracy. As seen in Table 2, constant scaling of 0.1 makes a statistically significant decrease in errors at the cost of increasing the time to complete task. However, the hypothesis does not appear to hold true when looking at the performance of constant scaling of 0.3 to 0.2 under delay. We believe this is since participants had so many errors, such as ring drops, when using constant scaling of 0.3, that it severely affected their time to complete the task.

By comparing positional scaling and constant scaling of 0.1 relative time to complete task performances under delay, it is evident that positional scaling did decrease the cost of time as intended via dynamic scaling. However, positional scaling did not retain the weighted error performance of constant scaling of 0.1 when under delay. We believe this occurs because our implementation of positional scaling does not count the slave-arms as obstacles. Therefore during the ring pass, which typically occurs in the center of the plot in Fig. \ref{fig:positionScalingPlot}, there is little to no additional scaling to improve the accuracy. The data from the user study supports this since 2 ring stretches and 7 ring drops during hand off were recorded for positional scaling under delay, and 0 ring stretches and 2 ring drops during hand off were recorded for constant scaling of 0.1 under delay. 

The final result is the performance of velocity scaling, which performed the best of all of the proposed solutions under delay. It has similar performance in weighted error as constant scaling of 0.1 at a lower cost of time. While under delay, velocity scaling on average reduced the weighted error rate by 43\% and 29\% and increased the time to complete task by no statistically significant margin and 22\% when compared against constant scaling of 0.3 and 0.2 respectively.

\section{Conclusion}

The results from the user study confirms with statistical significance our original hypothesis which states that decreasing the constant scaling factor will improve accuracy at the cost of time. Velocity scaling was shown to improve on this by achieving the same accuracy with a lower cost of time. Positional scaling did not perform as well, and it has the added challenge of requiring environmental information. Participants were also given little time to adjust to the proposed scaling solutions, so the performance improvements implies that they are intuitive. Furthermore, the simplicity of the constant scaling and velocity scaling solutions allows for them to be easily deployed on any teleoperational system under delay. Future work involves dynamically learning the appropriate scales for individual operators and delay conditions.

\section{Acknowledgements}
We thank all the participants from the user study for their time, all the members of the Advanced Robotics and Controls Lab at University of California San Diego for the intellectual discussions and technical help, and the UCSD Galvanizing Engineering in Medicine (GEM) program and the US Army AMEDD Advanced Medical Technology Initiative (AAMTI) for the funding.
\balance
\bibliographystyle{ieeetr}
\bibliography{references}

\end{document}